\documentclass{article}
\usepackage{spconf,amsmath,graphicx}
\usepackage{booktabs} 
\usepackage{float} 

\usepackage{enumitem}
\setlist{nosep, leftmargin=14pt}

\usepackage{mwe} 

\title{Spatial Sequence Attention Network for Schizophrenia Classification from Structural Brain MR Images}
\name{Nagur Shareef Shaik$^{*\dag}$, Teja Krishna Cherukuri$^{*\dag}$, Vince Calhoun$^{*\dag}$, Dong Hye Ye$^{*\dag}$
}
\address{$^*$Department of Computer Science, Georgia State University \\ 
 $^\dag$Translational Research in Neuroimaging and Data Science (TReNDS), Georgia State, Georgia Tech, Emory} 
\begin{document}
%
\maketitle
\begin{abstract}
Schizophrenia is a debilitating, chronic mental disorder that significantly impacts an individual's cognitive abilities, behavior, and social interactions. It is characterized by subtle morphological changes in the brain, particularly in the gray matter. These changes are often imperceptible through manual observation, demanding an automated approach to diagnosis. This study introduces a deep learning methodology for the classification of individuals with Schizophrenia. We achieve this by implementing a diversified attention mechanism known as Spatial Sequence Attention (SSA) which is designed to extract and emphasize significant feature representations from structural MRI (sMRI). Initially, we employ the transfer learning paradigm by leveraging pre-trained DenseNet to extract initial feature maps from the final convolutional block which contains morphological alterations associated with Schizophrenia. These features are further processed by the proposed SSA to capture and emphasize intricate spatial interactions and relationships across volumes within the brain. Our experimental studies conducted on a clinical dataset have revealed that the proposed attention mechanism outperforms the existing Squeeze \& Excitation Network for Schizophrenia classification.
\end{abstract}
\begin{keywords}
Schizophrenia (SZ), structural MRI (sMRI), Transfer Learning, Spatial Sequence Attention (SSA).
\end{keywords}

\section{Introduction}
\label{sec:introduction}

As per the World Health Organization (WHO), about 24 million people globally grapple with schizophrenia (SZ), a condition characterized by enduring hallucinations and disruptive behavior that profoundly affects their daily lives \cite{cannon2000prospective}. The symptoms of SZ can often overlap with those of other psychiatric disorders, making both diagnosis and treatment more challenging. Brain MRI data has proven valuable in uncovering consistent structural alterations in the brains of SZ patients, including reductions in hippocampal and thalamic volumes, along with increased globus pallidus volumes. However, these structural changes, while observable, provide only limited insights into the disorder \cite{haukvik2013schizophrenia}.

Advancements have been made in SZ classification in automated brain MRI analysis. Notably, researchers have explored methods to differentiate SZ from other psychiatric disorders and healthy controls based on gray matter density, employing support vector machine (SVM) models \cite{schnack2014can}. Additionally, some studies have used extended Granger causality to extract features from brain MRIs and subsequently selected features using Kendall's tau rank correlation coefficients, followed by SVM-based classification \cite{wismuller2021classification} \cite{wismuller2023large}. Recently, deep learning, with the convolutional neural network (CNN) architecture, has shown significant promise in the field of representation learning from images, even when dealing with 3D structural MRI (sMRI) scans. Prominent works have employed 3D CNNs for feature extraction to capture intricate spatial and morphological information in SZ patients \cite{sadeghi2022overview}. For instance, Hu et al. utilized a 3D CNN and pre-trained 2D CNN models to detect SZ using 3D sMRI data \cite{hu2022structural}. However, most of these works tried to use the holistic features learned by CNNs without using the attention mechanism to emphasize important features, which might yield performance improvement. 

The Squeeze and Excitation Network (SENet) is a neural network architecture with attention, designed to improve feature extraction and model relationships between channels in convolutional neural networks (CNNs) \cite{hu2018squeeze}. SENet dynamically recalibrates channel importance in feature maps based on spatial and channel-wise dependencies, enhancing the network's ability to capture fine details and improve performance in computer vision tasks, including medical image classification \cite{bodapati2021msenet}. However, SENet faces limitations in its ability to adapt effectively to higher-dimensional data and localization aspects \cite{shaik2022hinge}. To tackle this challenge, we develop a novel attention mechanism called Spatial Sequence Attention Network (SSANet). The SSANet's gating mechanism can mitigate and outperform challenges by selectively emphasizing SZ-specific information at spatial locations, learning to assign higher weights to relevant regions and focusing on crucial local structures in feature maps. Our experiments provide evidence supporting the superiority of our approach over SENet, especially in situations with constrained training data.

\begin{figure*}[ht]
    \centering
    \vspace{-10pt}
    \includegraphics[width=0.7\textwidth]{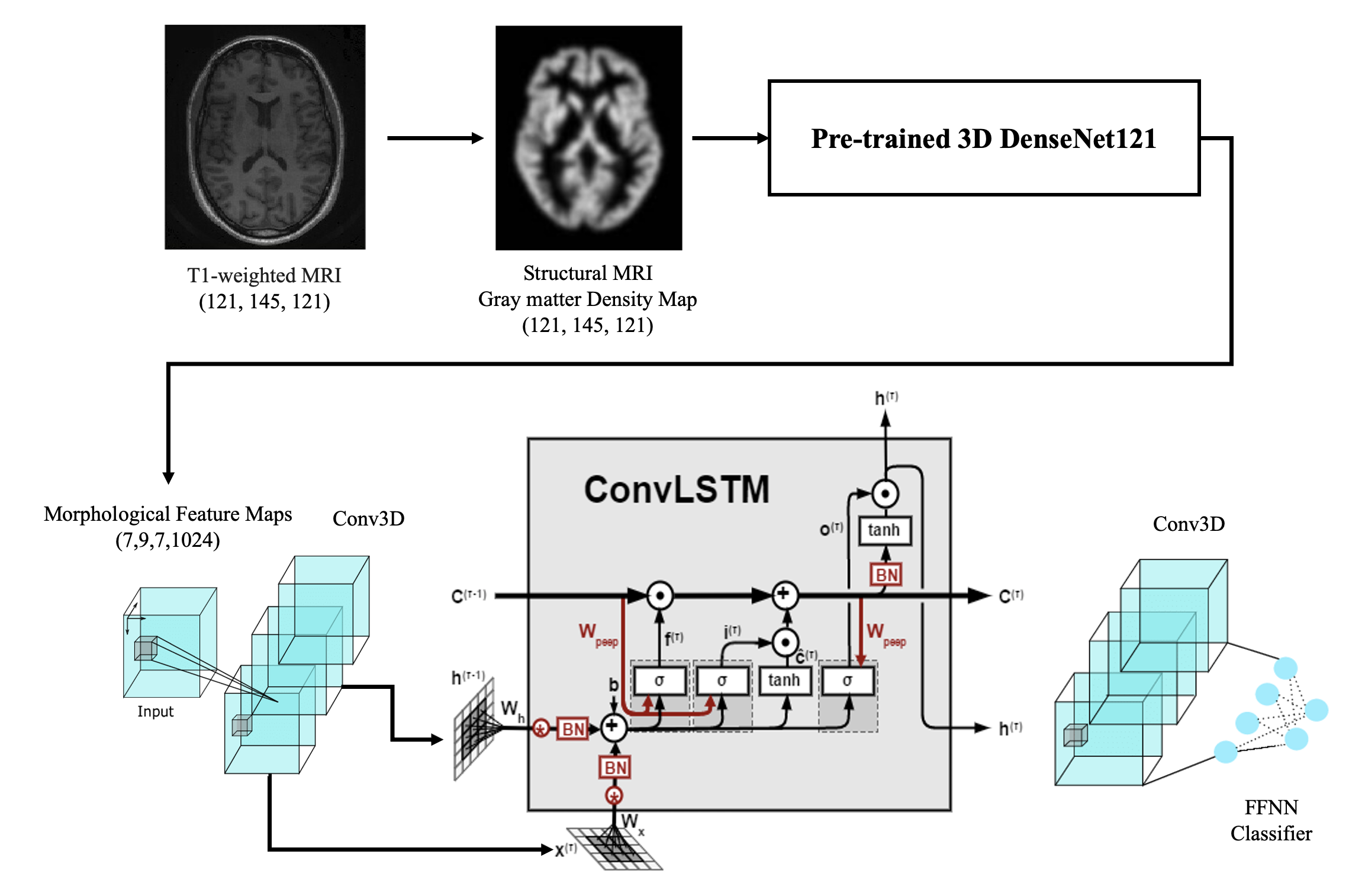}
    \vspace{-10pt}
    \caption{The proposed Spatial Sequence Attention Network (SSANet) architecture extracts morphological features from 3D T1-weighted MRI images (121x145x121 voxels) using a pre-trained 3D DenseNet121 model. The Spatial Sequence Attention (SSA) module then captures and accentuates complex spatial interactions and relationships within the brain volumes. Finally, a Feed Forward Neural Network (FFNN) classifies the images as schizophrenia or healthy.}
    \vspace{-10pt}
    \label{fig:proposed-architecture}
\end{figure*}

\vspace{-10pt}

\section{Methods}
\label{sec:methodology}

This study introduces the Spatial Sequence Attention Network (SSANet), a deep learning architecture for automated SZ classification from 3D T1-weighted MRI images with limited training data. Figure \ref{fig:proposed-architecture} illustrates the SSANet architecture. In the subsequent sections, we provide detailed technical insights into each phase of the proposed methodology.

\subsection{Morphological Representation Learning}

Over recent years, Convolutional Neural Networks (CNNs) have become prominent for learning representations from visual data. They employ a series of convolution operations to extract visual features by sliding a filter over the provided input image, capturing patterns and shapes through weighted summation of nearby pixel values, given the sufficient amount of training data \cite{lecun1995convolutional}. The availability of training data is scarce in the medical domain. To address this, Transfer Learning was widely employed to extract representation from a limited dataset using pre-trained models. Following the paradigm, we leverage a pre-trained DenseNet, comprising 121 layers, to extract morphological features, augmenting its capabilities through fine-tuning specifically on gray matter density sMRIs.

In Convolutional Neural Networks, the input from the previous layer $n-1$ is passed through a non-linear transformation denoted as $H_n$ to generate the input for the current layer $n$, represented as $X^{(n)} = H_n(X^{(n-1)})$. This direct back-to-front gradient propagation can lead to the issue of vanishing gradients in the network. Addressing this, Huang et. al. presented the DenseNet with the dense block architecture. The $n$-th layer establishes connections with all preceding $L$ layers, including the $n-1$-th layer, thereby facilitating improved gradient flow across the network \cite{huang2017densely}. Most of the previous works focus on extracting penultimate fully connected layer features. Apart from this, we utilize the convolutional feature maps. These features retain spatial and volumetric information essential for discriminating SZ. This type of transfer learning method proves advantageous and adept at meeting the criteria for extracting morphological features \cite{bodapati2021deep}.
\begin{equation}
    X_i = DenseNet121(I_i) \quad \forall i \in \{1, 2, ... N\}.
    \label{eq:inp_densenet}
\end{equation}
From the equation \ref{eq:inp_densenet}, the input gray matter density image for subject $i$ is denoted by $I_i$. It is a $(121 \times 145 \times 121)$ voxel image that is an input to a pre-trained 3D DenseNet121 model. The model first applies a 3D convolution with a $(7\times7\times7)$ kernel and pooling to the image. Then, it performs varying dense layers within each dense block. These layers include two 3D convolutions with different kernel sizes: $(1\times1\times1)$ and $(3\times 3 \times 3)$, which reduce the feature map size and improve computation. Three transition layers (ReLU, batch normalization, and dimension reduction pooling) connect the dense blocks. This produces a $(7 \times 9 \times 7 \times 1024)$ output vector $X_i$ for each subject $i$, which represents the morphological features of the subject's brain \cite{yu2022deep}.

\vspace{-10pt}

\subsection{Spatial Sequence Attention}
The Spatial Sequence Attention \cite{shaik2022hinge} is a special mechanism designed to capture spatial and volumetric dependencies in morphological feature maps $X_i$. It consists of three main components - a 3D convolutional (Conv3D) layer with 64 filters of a $(3\times 3 \times 3)$ kernel, a ConvLSTM layer with 64 filters having $(3\times 3 \times 3)$ contextual window, and a Conv3D layer with the same number of filters and kernel size as the input feature maps. Each component has its own significance in improving the provided feature maps. First Conv3D layer learns holistic features from the input. ConvLSTM is a variant of the LSTM (Long Short-Term Memory) with convolutional operations integrated into the cell \cite{NIPS2015_07563a3f}. It uses tanh and sigmoid activation functions to control the flow of information through the network and capture spatial information through sparse connections between the input and state transitions. This allows the model to learn intricate spatial interactions and relationships across volume in morphological feature maps. The significance of the current feature map is determined by considering both the current and past feature maps in the input volumetric sequence. The other Conv3D projects the output back to the original feature space.

High-level morphological feature representations obtained from the first Conv3D layer of SSA are processed through input gate ($I_t$), forget gate ($F_t$), and output gate ($O_t$), resulting in the evolution of the cell state ($\mathbf{C}_t$) and hidden state ($\mathbf{H}_t$) at a specific time step $t$. The future state of a specific feature map is determined by considering the inputs and previous states of neighboring feature maps within the receptive field. Convolutional operations in the transitions from input to state and from state to state facilitate this process. The ConvLSTM cell operates based on a set of fundamental equations involving convolution ($*$) and element-wise product ($\odot$). Resultants are attention-based feature representations that are subsequently passed as input to a regularized fully connected neural network to evaluate the presence of SZ in a given sMRI image.
\begin{equation}
I_t = \sigma(W_{xi} * \mathbf{X}_t + W_{hi} * \mathbf{H}_{t-1} + W_{ci} * \mathbf{C}_{t-1} + \mathbf{b}_i) 
\end{equation}
\begin{equation}
F_t = \sigma(W_{xf} * \mathbf{X}_t + W_{hf} * \mathbf{H}_{t-1} + W_{cf} * \mathbf{C}_{t-1} + \mathbf{b}_f)
\end{equation}
\begin{equation}
    \mathbf{C}_{t} = F_t \odot \mathbf{C}_{t} + I_t \odot \tanh(W_{xc} * \mathbf{X}_t + W_{hc} * \mathbf{H}_{t-1} + \mathbf{b}_c)
\end{equation}
\begin{equation}
    O_t = \sigma(W_{xo} * \mathbf{X}_t + W_{ho} * \mathbf{H}_{t-1} + W_{co} * \mathbf{C}_{t-1} + \mathbf{b}_o) \mathbf{H}_{t} 
\end{equation}

The above set of equations represents crucial steps in processing morphological feature maps while capturing spatial dependencies. The input gate ($I_t$) assesses the importance of each input feature map in sequence ($\mathbf{X}_t$), for inclusion in the cell state ($\mathbf{C}_{t}$). The forget gate ($F_t$) determines the extent to which features from the previous cell state ($\mathbf{C}_{t-1}$) should be retained. The cell state ($\mathbf{C}_{t}$) is then updated by selectively integrating features from the input gate and forgetting unwanted features. Subsequently, the output gate ($O_t$) regulates the exposure of the updated cell state as the output ($\mathbf{H}_{t}$). The hidden state ($\mathbf{H}_{t}$) is derived by applying the output gate to the hyperbolic tangent of the updated cell state, ensuring that the ConvLSTM captures and utilizes both short and long-term dependencies within sequential data, making it particularly effective for tasks involving spatial and volumetric information. The weight matrices ($W_{xi}, W_{hi}, W_{ci}, \ldots$) and bias vectors ($\mathbf{b}_i, \mathbf{b}_f, \mathbf{b}_c, \ldots$) associated with each gate are learned during the training process, enabling the ConvLSTM to adapt to the specific characteristics of the data it processes.

\vspace{-10pt}

\subsection{Regularized Feed Forward Neural Network} 
The Feed Forward Neural Network (FFNN) comprises two dense layers, with 512 and 256 neurons, employing the Gaussian Error Linear Unit (GELU) activation function, and includes a softmax layer for class labeling. To mitigate overfitting, each dense layer is accompanied by a dropout layer set at a rate of 0.5. Furthermore, weights in the fully connected layers are subjected to $L_1$ regularization, while biases undergo $L_1L_2$ regularization, providing additional protection against overfitting during the training process. These three components are stacked and trained end-to-end through gradient centralization to make predictions, collectively referred to as the Spatial Sequence Attention Network. The combination of induced dropout, parameter regularization, and gradient centralization serves to prevent overfitting, allowing the model to generate meaningful predictions with reasonable accuracy. However, it is worth noting that the model's performance can vary depending on the pre-trained convolutional base used for morphological feature extraction.

\vspace{-10pt}

\section{Experiments}
\label{sec:experiments}

In this section, we start by presenting the details of the clinical dataset employed in our experiments. Subsequently, we describe the experimental setup, evaluation metrics, and conclude with both quantitative and qualitative analyses of the results.

\vspace{-10pt}

\subsection{Dataset}

To validate the effectiveness of the proposed Spatial Sequence Attention network, we utilized a portion of the FBIRN dataset, which comprises both chronic SZ patients and control individuals \cite{keator2016function}. This dataset encompasses various data modalities, including structural MRI (sMRI), functional MRI (fMRI), single-nucleotide polymorphisms (SNP), and patient behavior information. However, we used just sMRI for our experiments to validate the effectiveness of the attention mechanism on morphological features. In total, this dataset consists of 186 participants, with 82 individuals diagnosed with SZ and 104 healthy subjects. Each MRI scan is downsampled by dimensions (121, 145, 121) for efficient processing. 

During experiments on this dataset, various hyperparameter values were explored, including learning rates (0.005 to 0.05), dropout rates (0.1 to 0.5), and regularization rates (0.001 to 0.1). Selected values to optimize model performance: initial learning rate (0.0001), batch size (32), training epochs (100), with L2 weight regularization (0.005) and L1\_L2 bias regularization (0.005).

\subsection{Quantitative Evaluation}
For quantitative evaluation, we employ 5-fold cross-validation for SZ classification on the FBIRN dataset. We use standard classification metrics such as Accuracy (Acc), Precision (Prec), Recall, and $F_1$ Score to report the performance of conducted experiments. As a baseline model for our attention mechanism, we incorporate the existing Squeeze and Excitation Attention (SENet) into the feature extraction process of pre-trained DenseNet121 for the classification of SZ based on morphological MRI features. Subsequently, we aim to assess the effectiveness of the proposed Spatial Sequence Attention (SSANet) in terms of the feature extraction process of pre-trained DenseNet121. Additionally, we provide a comparative analysis with recent studies in the field that have utilized the same dataset and experimental configuration \cite{yu2022deep} \cite{kanyal2023multi}, presenting the results in Table \ref{tab:results}.

\begin{table}[H]
\centering
\caption{Comparison of Five-fold Cross-validation Metrics (in \%) for SZ Classification}
\label{tab:results}
\begin{tabular}{lcccc}
\toprule
Method & Acc & Prec & Recall & $F_1$ \\
\midrule
DenseNet \cite{huang2017densely} & 70.00 & - & - & - \\
XGBoost \cite{kanyal2023multi} & 71.51 & - & - & - \\
SENet & 72.80 & 77.84 & 68.06 & 66.79 \\
& $\pm$ 0.05 & $\pm$ 0.06 & $\pm$ 0.08 & $\pm$ 0.09 \\
SSANet (Ours) & 75.12 & 77.22 & 72.76 & 71.91 \\
& $\pm$ 0.06 & $\pm$ 0.04 & $\pm$ 0.08 & $\pm$ 0.09 \\
\bottomrule
\end{tabular}
\end{table}

Looking at Table \ref{tab:results}, it is evident that our proposed SSANet outperformed not only the Squeeze \& Excitation Network (SENet) but also the fine-tuned DenseNet \cite{huang2017densely} and DenseNet features combined with XGBoost \cite{kanyal2023multi} in terms of accuracy. SENet, on the other hand, delivered competitive results. SSANet achieved higher accuracy and more balanced performance metrics. Additionally, the standard deviations accompanying our results highlight the stability of SSANet's performance. These findings underscore the efficacy of our proposed model in enhancing the SZ classification based on morphological sMRI features and emphasize its potential as a valuable addition to the field of medical image analysis and diagnostics.

\vspace{-10pt}

\subsection{Qualitative Evaluation}
\begin{figure}[H]
    \centering
    \vspace{-10pt}
    \includegraphics[width=0.45\textwidth]{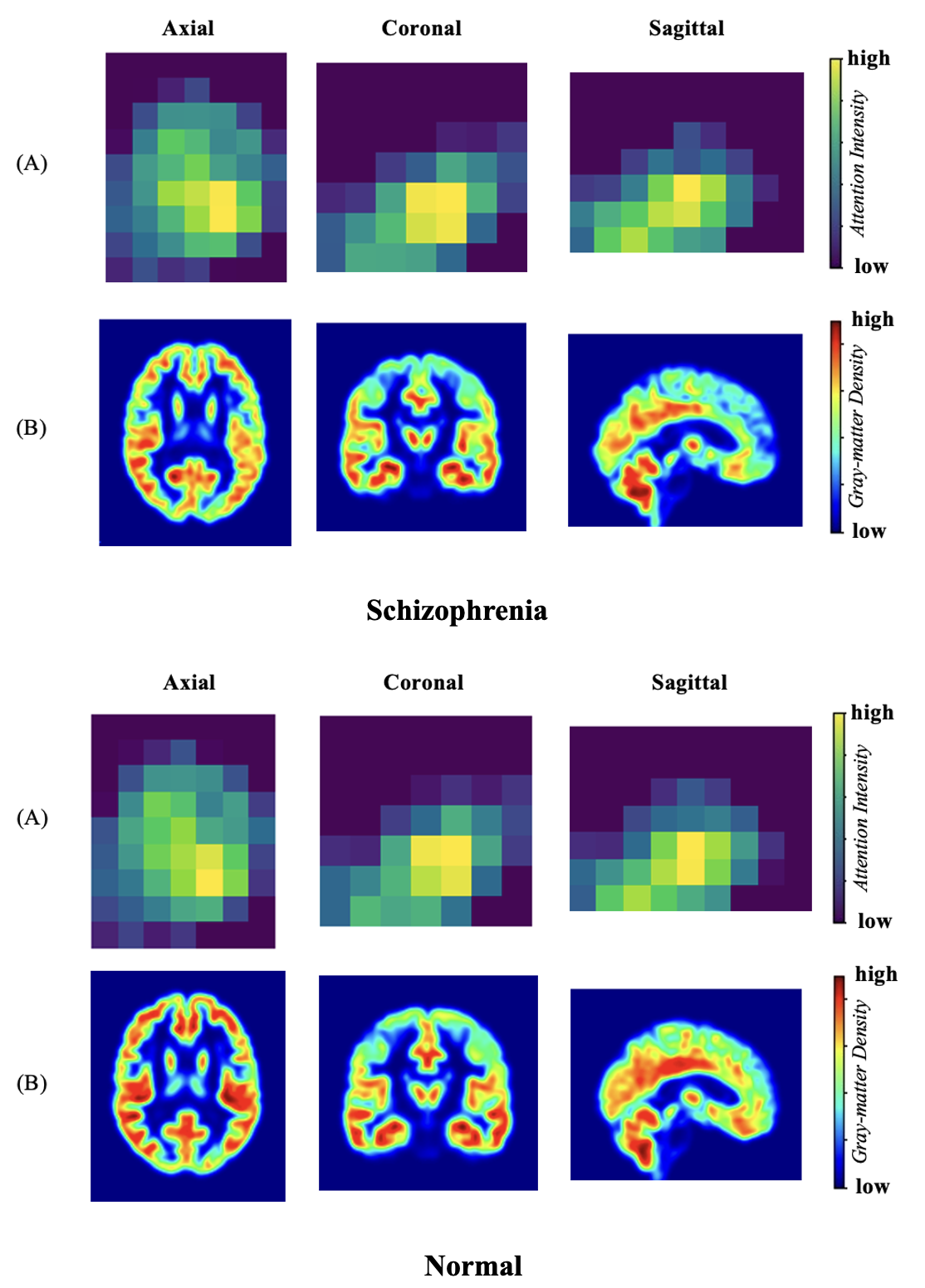}
    \caption{Visualization of (A) Attention Heatmaps; (B) Gray Matter Density Maps of a SZ (Top) and a normal (Bottom) subject; SSANet correctly gives attention around the disease impacted areas such as frontal lobe and temporal lobe. Note the decreased gray matter density in the SZ subject compared with the normal subject in those regions.}
    \vspace{-10pt}
    \label{fig:attention-maps}
\end{figure}

The qualitative evaluation involves visualizing attention heatmaps for axial, coronal, and sagittal slices. The superior performance of our proposed method for SZ classification is evident from the attention heatmaps from Figure \ref{fig:attention-maps}, which can identify brain regions (frontal lobe, and temporal lobe) for attention and capture the gray matter density loss in SZ subjects. These attention maps contribute to enhanced discriminatory features, showcasing the effectiveness of our approach in capturing nuanced abnormalities associated with schizophrenia.
\section{Conclusion}
\label{sec:conclusion}
In this paper, we introduced the Spatial Sequence Attention Network (SSANet), a novel deep-learning model addressing automated SZ classification with limited training data. SSANet successfully captures intricate spatial interactions and relationships across volumes within the brain, yielding improved SZ classification performance compared with the Squeeze \& Excitation Network (SENet) and DenseNet without attention mechanism. This highlights its potential for enhancing schizophrenia diagnosis despite limited training data and complex spatial dependencies. As a future work, our model's interpretability and the clinical relevance of its learned features will be further investigated. Additionally, we will explore aspects of our proposed method, emphasizing multi-modal data integration, interpretability, ethical considerations, and generalization across demographics. These directions hold the potential for enhancing diagnostic accuracy and ensuring equitable healthcare.

\bibliographystyle{IEEEbib}
\bibliography{refs}

\begin{thebibliography}{10}

\bibitem{cannon2000prospective}
Tyrone~D Cannon, Isabelle~M Rosso, J~Megginson Hollister, Carrie~E Bearden, Laura~E Sanchez, and Trevor Hadley,
\newblock ``A prospective cohort study of genetic and perinatal influences in the etiology of schizophrenia,''
\newblock {\em Schizophrenia bulletin}, vol. 26, no. 2, pp. 351--366, 2000.

\bibitem{haukvik2013schizophrenia}
Unn~Kristin Haukvik, Cecilie~Bhandari Hartberg, and Ingrid Agartz,
\newblock ``Schizophrenia--what does structural mri show?,''
\newblock {\em Tidsskrift for Den norske legeforening}, 2013.

\bibitem{schnack2014can}
Hugo~G Schnack, Mireille Nieuwenhuis, Neeltje~EM van Haren, Lucija Abramovic, Thomas~W Scheewe, Rachel~M Brouwer, Hilleke E~Hulshoff Pol, and Ren{\'e}~S Kahn,
\newblock ``Can structural mri aid in clinical classification? a machine learning study in two independent samples of patients with schizophrenia, bipolar disorder and healthy subjects,''
\newblock {\em Neuroimage}, vol. 84, pp. 299--306, 2014.

\bibitem{wismuller2021classification}
Axel Wism{\"u}ller and M~Ali Vosoughi,
\newblock ``Classification of schizophrenia from functional mri using large-scale extended granger causality,''
\newblock in {\em Medical Imaging 2021: Computer-Aided Diagnosis}. SPIE, 2021, vol. 11597, pp. 337--350.

\bibitem{wismuller2023large}
Axel Wism{\"u}ller, Ali Vosoughi, Akhil Kasturi, and Nathan Hadjiyski,
\newblock ``Large-scale augmented granger causality (lsagc) for discovery of causal brain connectivity networks in schizophrenia patients using functional mri neuroimaging,''
\newblock in {\em Medical Imaging 2023: Biomedical Applications in Molecular, Structural, and Functional Imaging}. SPIE, 2023, vol. 12468, pp. 278--286.

\bibitem{sadeghi2022overview}
Delaram Sadeghi, Afshin Shoeibi, Navid Ghassemi, Parisa Moridian, Ali Khadem, Roohallah Alizadehsani, Mohammad Teshnehlab, Juan~M Gorriz, Fahime Khozeimeh, Yu-Dong Zhang, et~al.,
\newblock ``An overview of artificial intelligence techniques for diagnosis of schizophrenia based on magnetic resonance imaging modalities: Methods, challenges, and future works,''
\newblock {\em Computers in Biology and Medicine}, vol. 146, pp. 105554, 2022.

\bibitem{hu2022structural}
Mengjiao Hu, Xing Qian, Siwei Liu, Amelia~Jialing Koh, Kang Sim, Xudong Jiang, Cuntai Guan, and Juan~Helen Zhou,
\newblock ``Structural and diffusion mri based schizophrenia classification using 2d pretrained and 3d naive convolutional neural networks,''
\newblock {\em Schizophrenia research}, vol. 243, pp. 330--341, 2022.

\bibitem{hu2018squeeze}
Jie Hu, Li~Shen, and Gang Sun,
\newblock ``Squeeze-and-excitation networks,''
\newblock in {\em Proceedings of the IEEE conference on computer vision and pattern recognition}, 2018, pp. 7132--7141.

\bibitem{bodapati2021msenet}
Jyostna~Devi Bodapati, Shaik~Nagur Shareef, Veeranjaneyulu Naralasetti, and Nirupama~Bhat Mundukur,
\newblock ``Msenet: Multi-modal squeeze-and-excitation network for brain tumor severity prediction,''
\newblock {\em International Journal of Pattern Recognition and Artificial Intelligence}, vol. 35, no. 07, pp. 2157005, 2021.

\bibitem{shaik2022hinge}
Nagur~Shareef Shaik and Teja~Krishna Cherukuri,
\newblock ``Hinge attention network: A joint model for diabetic retinopathy severity grading,''
\newblock {\em Applied Intelligence}, vol. 52, no. 13, pp. 15105--15121, 2022.

\bibitem{lecun1995convolutional}
Yann LeCun, Yoshua Bengio, et~al.,
\newblock ``Convolutional networks for images, speech, and time series,''
\newblock {\em The handbook of brain theory and neural networks}, vol. 3361, no. 10, pp. 1995, 1995.

\bibitem{huang2017densely}
Gao Huang, Zhuang Liu, Laurens Van Der~Maaten, and Kilian~Q Weinberger,
\newblock ``Densely connected convolutional networks,''
\newblock in {\em Proceedings of the IEEE conference on computer vision and pattern recognition}, 2017, pp. 4700--4708.

\bibitem{bodapati2021deep}
Jyostna~Devi Bodapati, Nagur~Shareef Shaik, and Veeranjaneyulu Naralasetti,
\newblock ``Deep convolution feature aggregation: an application to diabetic retinopathy severity level prediction,''
\newblock {\em Signal, Image and Video Processing}, vol. 15, pp. 923--930, 2021.

\bibitem{yu2022deep}
H.~Yu, T.~Florian, V.~Calhoun, and D.~Ye,
\newblock ``Deep learning from imaging genetics for schizophrenia classification,''
\newblock in {\em 2022 IEEE International Conference on Image Processing (ICIP)}. IEEE, 2022, pp. 3291--3295.

\bibitem{NIPS2015_07563a3f}
Xingjian SHI, Zhourong Chen, Hao Wang, Dit-Yan Yeung, Wai-kin Wong, and Wang-chun WOO,
\newblock ``Convolutional lstm network: A machine learning approach for precipitation nowcasting,''
\newblock in {\em Advances in Neural Information Processing Systems}. 2015, vol.~28, pp. 802--810, Curran Associates, Inc.

\bibitem{keator2016function}
David~B Keator, Theo~GM van Erp, Jessica~A Turner, Gary~H Glover, Bryon~A Mueller, Thomas~T Liu, James~T Voyvodic, Jerod Rasmussen, Vince~D Calhoun, Hyo~Jong Lee, et~al.,
\newblock ``The function biomedical informatics research network data repository,''
\newblock {\em Neuroimage}, vol. 124, pp. 1074--1079, 2016.

\bibitem{kanyal2023multi}
Ayush Kanyal, Srinivas Kandula, Vince Calhoun, and Dong~Hye Ye,
\newblock ``Multi-modal deep learning on imaging genetics for schizophrenia classification,''
\newblock in {\em 2023 IEEE International Conference on Acoustics, Speech, and Signal Processing Workshops (ICASSPW)}. IEEE, 2023, pp. 1--5.

\end{thebibliography}

\end{document}